\pdfoutput=1

\documentclass[11pt]{article}

\usepackage[]{acl}

\usepackage{times}
\usepackage{latexsym}
\usepackage{multirow}
\usepackage[T1]{fontenc}

\usepackage[utf8]{inputenc}
\usepackage{algorithm2e}
\usepackage{enumitem}

\usepackage{microtype}
\usepackage{graphicx}
\usepackage{amsmath}
\usepackage{subcaption}
\usepackage{xcolor}
\usepackage{booktabs}
\usepackage{multirow}
\usepackage{todonotes}

\title{A Study on Prompt-based Few-Shot Learning Methods for Belief State Tracking in Task-oriented Dialog Systems}

\author {
        Debjoy Saha ,
        Bishal Santra ,
        Pawan Goyal \\
        Indian Institute of Technology, Kharagpur, India \\
        sahadebjoy10@iitkgp.ac.in, bsantraigi@gmail.com, pawang@cse.iitkgp.ac.in}

\begin{document}
\maketitle
\begin{abstract}
We tackle the Dialogue Belief State Tracking \emph{(DST)} problem of task-oriented conversational systems. Recent approaches to this problem leveraging Transformer-based models have yielded great results. However, training these models is expensive, both in terms of computational resources and time. Additionally, collecting high quality annotated dialogue datasets remains a challenge for researchers because of the extensive annotation requirement.
Driven by the recent success of pre-trained language models and prompt-based learning, we explore prompt-based few-shot learning for Dialogue Belief State Tracking. We formulate the DST problem as a 2-stage prompt-based language modelling task and train language models for both tasks and present a comprehensive empirical analysis on their separate and joint performance. We demonstrate the potential of prompt-based methods in few-shot learning for DST and provides directions for future improvement.
\end{abstract}

\section{Introduction}

Dialogue Belief State Tracking is a central problem for task-based conversational systems. The Belief State maintains a distribution of states across different dialogue turns summarising the conversation state at any point by extracting the intent from the user and system inputs. The belief state is used by the system to take appropriate actions at each turn until the conversation is concluded and the user goal is achieved. 

Recent State-of-the-art models that tackle the Belief State Tracking problem are generally based on large language models \citep{hosseini2020simple,heck2020trippy,wu2019transferable}. Their training usually involves huge computation and data requirements, one or both of which might be unavailable. The development of models like BERT \citep{devlin2018bert} and GPT-2 \citep{Radford2019LanguageMA} has also inspired advances in the use of pre-trained language models (\emph{PLMs}) for low-resource few-shot learning for dialog generation \citep{Zhao2020Low-Resource}. The recent paradigm of \emph{prompt-based learning} \citep{brown2020language} equips PLMs with constructive task-dependent prompts to simplify language generation for downstream tasks. This method has shown great results on few-shot and zero-shot learning tasks such as classification \citep{gao-etal-2021-making,schick-schutze-2021-exploiting,han2021ptr} and text generation \citep{schick2020few,li-liang-2021-prefix}. Relatively fewer attempts have been made towards few-shot learning for the DST task of dialog systems. \newcite{dingliwal2021few} presents a few-shot meta-learning approach to DST. \newcite{Peng2020SOLOISTFT, madotto2021few} show few-shot DST performance on just single domain subsets of data. \newcite{Madotto2020LanguageMA} shows few-shot training results on the speech \emph{ACT}(Active Intent) identification task of task-oriented datasets. However, none of the papers present baselines on the end-to-end multi-domain, multi-slot belief-state tracking. 

In this paper, we make the first step towards evaluating prompt-based few-shot learning for the end-to-end dialogue to belief state prediction task. Specifically, we formulate the DST task as a two stage language generation problem and provide few-shot performance using pretrained language models like GPT-2, BERT and T5. Our analysis questions the viability of tackling DST in a prompt-based few-shot setting.

\begin{figure*}[h]
    \centering
    \includegraphics[width=0.65\textwidth]{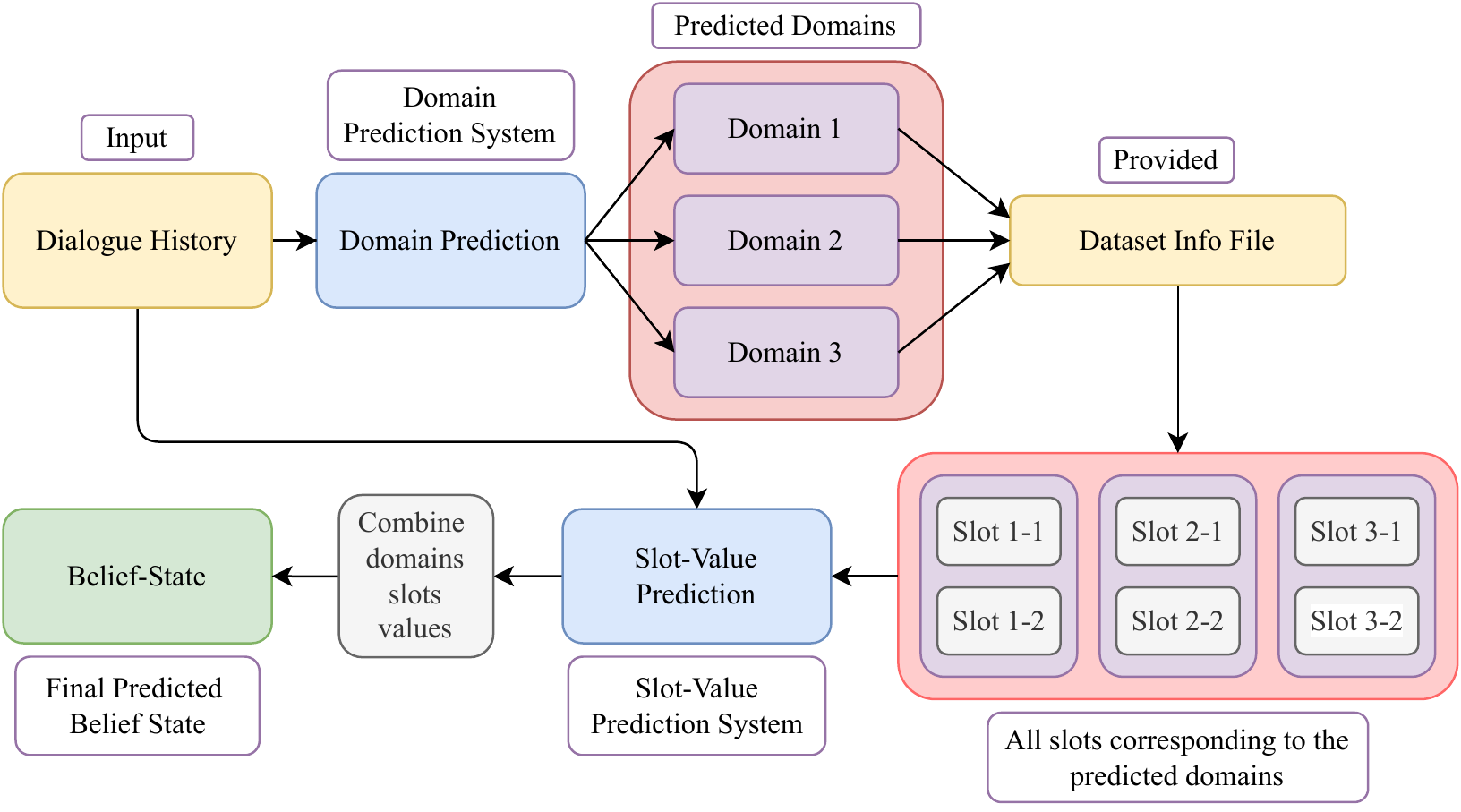}
    \caption{The Dialogue History to Belief State Pipeline}
    \label{fig:pipeline1}
\end{figure*}

\section{Proposed Method}
\paragraph{A prompt-based DST pipeline:}
We present a top-level overview of the belief-state prediction pipeline in Figure \ref{fig:pipeline1}. The first part of the pipeline, the \textbf{domain prediction system}, predicts a set of domains, and the second part of the pipeline, \textbf{slot prediction system}, assigns values to each of the slots belonging to these domains and constructs the entire belief-state.

\subsection{Domain Prediction} \label{domainpred}
In this part of the pipeline, the task is to predict one or multiple domains from the input text comprising the dialogue history and the prompt. We adopt two different approaches for this.

\noindent\textbf{Using Masked Language Modelling:} We extend the dialogue history by the appropriate prompt, containing up to 4 masks, which corresponds to the maximum number of domains that are referred to for any example. For each mask, we select the domain with the highest language modelling score. 


\noindent\textbf{Masked Prompt Design:}
The key challenge in designing the prompt is to have a phrase that can be applied to all the domains normally, thus achieving minimal zero-shot perplexity and maximum few-shot learnability. We adopt the prompt ``Excited to see the \emph{[MASK]}.'' and its multi-mask variants, based on their zero-shot performances, among other similar prompts, like this 2-mask variant ``Thank you for the information on the \emph{[MASK]} and \emph{[MASK]}''. We also restrict the vocab for domain prediction using MLM to help the model to avoid wrong predictions.

\noindent\textbf{Masked Language Modelling Inference:} 
An important limitation of MLM-based domain prediction is the inability to know the number of domains at run-time, which prevents us from directly using the appropriate prompt for generation. To address this issue, we propose \textbf{Weighted Grouped Token Scores (WGS)}. In this method, the model is run four times, once with each prompt to get a set of domain predictions $D_k$ (each containing $k$ prompts) for $k \in [1,4]$, and the averaged scores are normalised by a weighting factor $w_k$, to minimise the effects induced by changing the number of masks. The score can be expressed by the equation: $S_k = \frac{1}{k \times w_k} \times \sum_{d \in D_k} q(d)$, where $q(d)$ is the softmax score for domain $d \in D_k$.
The final set of predicted tokens comes out as $\hat D = argmax_k(S_k)$. The weights $w_k$ are learnt using a \textbf{Genetic Algorithm}\footnote{ Genetic algorithm \cite{538609} is a type of evolutionary computer algorithm that can be used to determine good solutions to optimisation problems.}, over the training set. 
It was observed that the genetic algorithm assigned almost the same weights to 1, 2, and 3-domain predictions (around 0.35) and a significantly higher weight (0.8) to the 4-domain predictions, indicating that the trained model over-incentivizes 4-domain predictions. However, these weights can be expected to change based on the dataset used.

\noindent\textbf{Using Causal Language Modelling:} Using causal language modelling can help remove the issues arising during masked-language model inference, as the generation process is unconstrained and can predict variable number of tokens. We extend the dialogue history by the appropriate prompt, and train the model to predict the appropriate domain string containing all the domains. An issue with making inferences from models trained using datasets whose majority of samples are for 1 or 2 domains was that the output was frequently cut-short to two or fewer domains. To predict longer sequences, we adopt \textit{Unlikelihood Training} for \emph{EOS} (End-of-Sentence)-tokens, similar to \newcite{welleck2019neural}. 


\noindent\textbf{Causal Prompt Design:} 
We select a \emph{QA}-style prompt \textit{``What are the mentioned domains?''} as, (a)~Prompts similar to the one considered are used in the demo examples provided in \newcite{brown2020language} for GPT-2, and (b)~\emph{QA}-style prompts for GPT-2 showed better zero-shot performance as compared to \emph{continuation}-prompts.


\noindent\textbf{Evaluation metrics:}
For evaluating domain predictions, we used the metric \textbf{Full Accuracy} \emph{(FA)}. For ground-truth set of domains $D_{gold}$ and predicted set of domains $D_{pred}$, the accuracy is given by the equation: $FA = (D_{pred} == D_{gold})$.

\subsection{Slot-Value Predictions} \label{slotpred}
In this sub-problem, we generate values for each slot which is a part of the identified domains. Since these predictions need not necessarily conform to a predetermined template, we just adopt the causal language modelling approach here.

\noindent\textbf{Causal LM Training:} This task is formulated similar to the causal LM based domain prediction, the only difference being that we use slot-specific prompts. For example, for \emph{hotel-name}, the prompt can be \textit{``What is the name of the hotel?''}.

\noindent\textbf{Evaluation metrics:} In addition to accuracy, we define a flexible accuracy measure which allows small mistakes in the generated outputs, for example, capitalisation or punctuation.

\section{Experimental Setup}
\noindent\textbf{Dataset:} We use the \textbf{MultiWOZ-2.2} dataset \citep{zang2020multiwoz}, a large-scale, multi-domain dialogue dataset of human-human conversations. Dialogues span over eight domains (restaurant, train, attraction, hotel, taxi, hospital, police, bus) and over 61 domain-slot pairs (\emph{hotel-name}, \emph{hotel-type}, \emph{train-arriveby} and so on). We sample ideally distributed datasets for both the problems. 

\noindent\textbf{Implementation Details:} For masked domain-prediction, we used pretrained masked language model BERT \citep{devlin2018bert}. We use GPT-2 \citep{brown2020language} for CLM-based domain prediction. For the slot-value prediction task, we use GPT-2 \cite{Radford2019LanguageMA}, GPT-neo \cite{brown2020language} and T5 \cite{raffel2019exploring}. A detailed description of the hyper-parameters and computational resources are provided in the appendix.

\section{Results and Analysis} \label{res}

\subsection{Domain Prediction} \label{sec:BERTM}
\begin{figure}[!ht]
    \centering
    \begin{subfigure}[b]{0.9\linewidth}
        \centering
        \includegraphics[width=\linewidth]{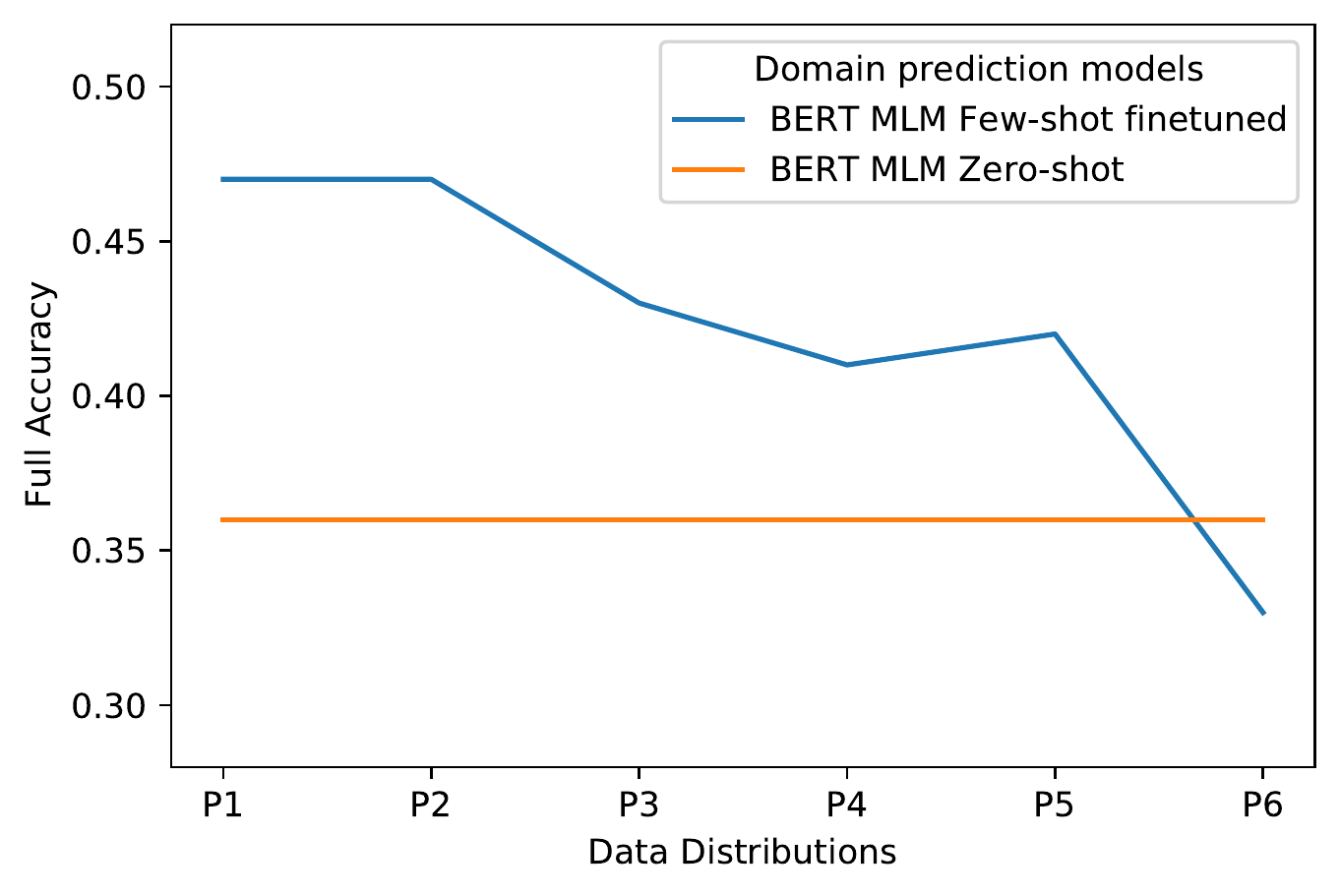}
        \caption{Domain prediction full accuracy variation with data distribution for BERT-MLM}
        \label{fig:FA}
    \end{subfigure}
    \begin{subfigure}[b]{0.9\linewidth}
        \centering
        \includegraphics[width=\linewidth]{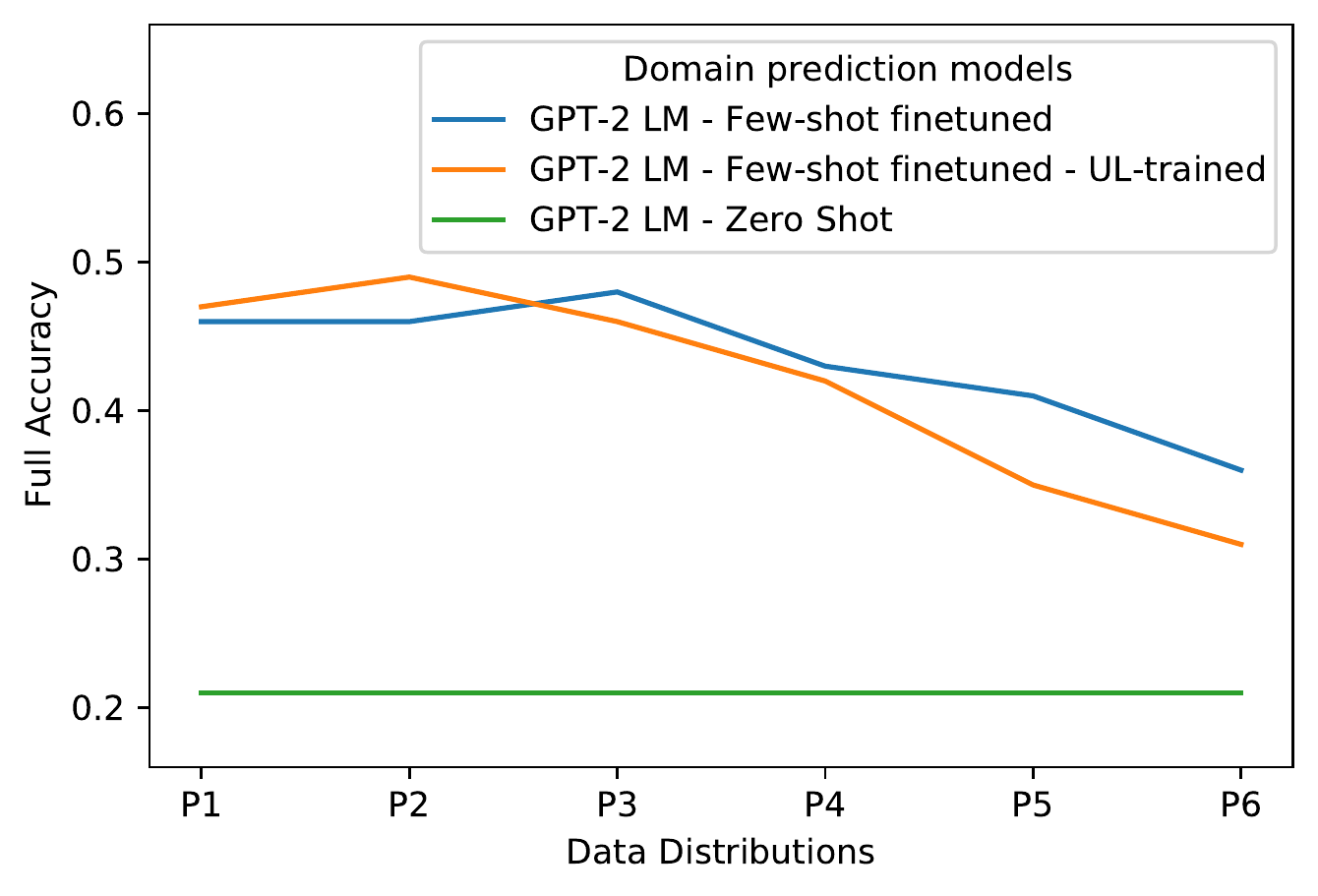}
        \caption{Domain prediction full accuracy variation with data distribution for GPT2-LM}
        \label{fig:FA2}
    \end{subfigure}
    \begin{subfigure}[b]{0.9\linewidth}
        \centering
        \includegraphics[width=\linewidth]{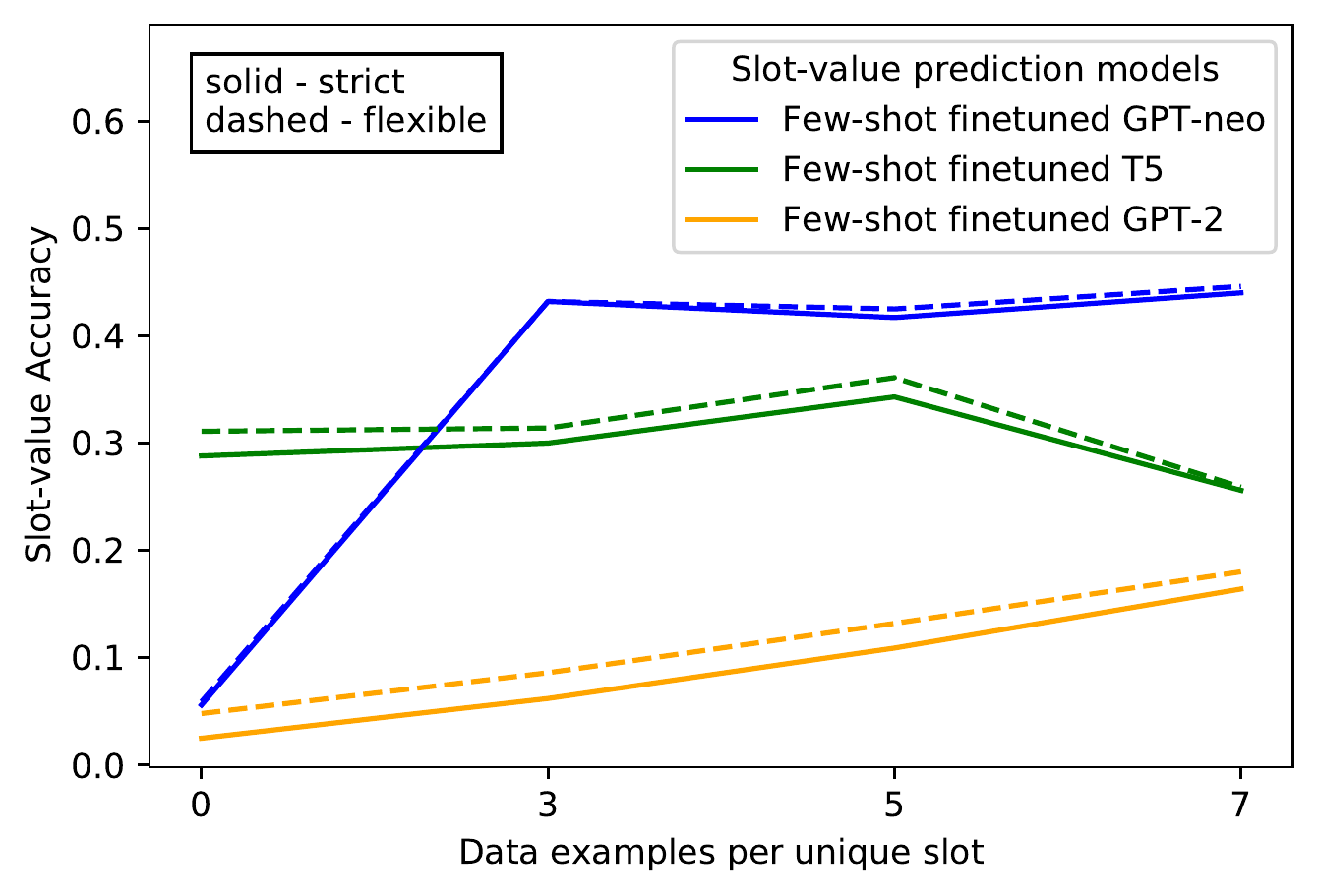}
        \caption{Slot-value prediction accuracy comparison of GPT-neo, GPT2 and T5 with dataset size}
        \label{fig:strict}
    \end{subfigure}
    \caption{Results from the two stages in our pipeline.}
\end{figure}

The \textbf{BERT-MLM} model was trained on datasets containing \textbf{128} training examples with different data distributions, containing different proportions of 1, 2, 3 and 4-domain data points. We have shown results for various sample distributions in Figure \ref{fig:FA}, the proportion of 3 and 4-domain data-points in the training data is increased from left to right (P1 to P6).

The proportion of test-data having just 1 or 2 ground-truth domains was found to be much higher. Consequently, the full accuracy (Figure \ref{fig:FA}) metric shows a gradual decline as the proportion of such data reduces in the training dataset. The overall best performance was obtained as a full accuracy of \textbf{0.47} 
with the data proportions 0.35, 0.35, 0.15 and 0.15 for 1, 2, 3, and 4-domain data respectively.

The \textbf{GPT2-LM} model was run on datasets with the same data distributions and dataset size (=128) as BERT-MLM. We present the accuracy metric full accuracy in Figure \ref{fig:FA2} for models trained (a) without and (b) with the unlikelihood loss component. The unlikelihood-trained models slightly outperform in a less-domain data setting only to get much worse as the proportion of data with large number of ground-truth domains increases. The reason for this is the over-prediction of domains in the latter case. The best metrics are obtained for the same data distribution as full accuracy of \textbf{0.49} for the unlikelihood-trained model.

\subsection{Slot-value Prediction} 
The slot-value prediction models were trained on datasets containing a fixed amount of data samples per-slot. The prediction accuracy is displayed in Figure \ref{fig:strict} (with the 0-data-per-slot label referring to the zero-shot accuracy). GPT-neo demonstrates the best performance, reaching up to \textbf{0.44 accuracy} with 7 data-points per sample (which amounts to a total data-size of 125), followed by T5 with \textbf{0.34 accuracy} with 5 data-points per sample. More detailed results are presented in the appendix.



\subsection{Full Belief-state Prediction} 
For getting the final predictions, we use both the best performing domain-prediction models (BERT-MLM and GPT2-LM trained using unlikelihood loss) trained on the 128-sized dataset and the T5-based slot-value prediction model, trained using a dataset containing 5 data points per unique slot giving a total of 80 training samples. T5 is preferred over GPT-neo for slot-prediction, as GPT-neo gives low combined performance due to over-prediction of values (predicting the same value for a set of similar slots) and was observed to generate less slots on average than the T5 model. The final metrics come out as \textbf{Joint Accuracy = 1\%} and \textbf{Slot Accuracy = 56\%}

\begin{figure}[!htbp]
    \centering
    \begin{subfigure}[b]{0.9\linewidth}
        \centering
        \includegraphics[width=\linewidth]{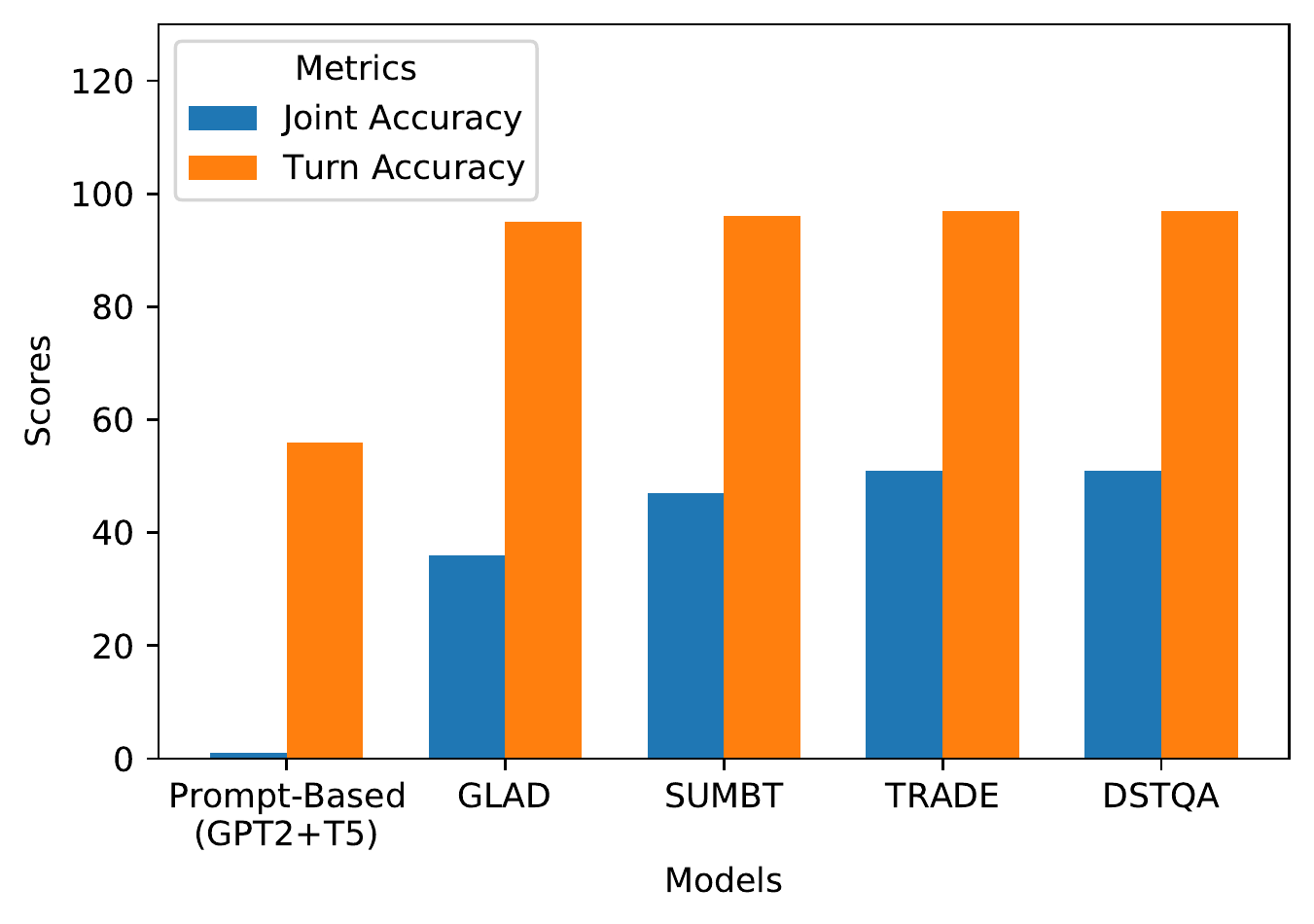}
        \caption{End-to-end trained models}
        \label{fig:model}
    \end{subfigure}
    \begin{subfigure}[b]{0.9\linewidth}  
        \centering 
        \includegraphics[width=\linewidth]{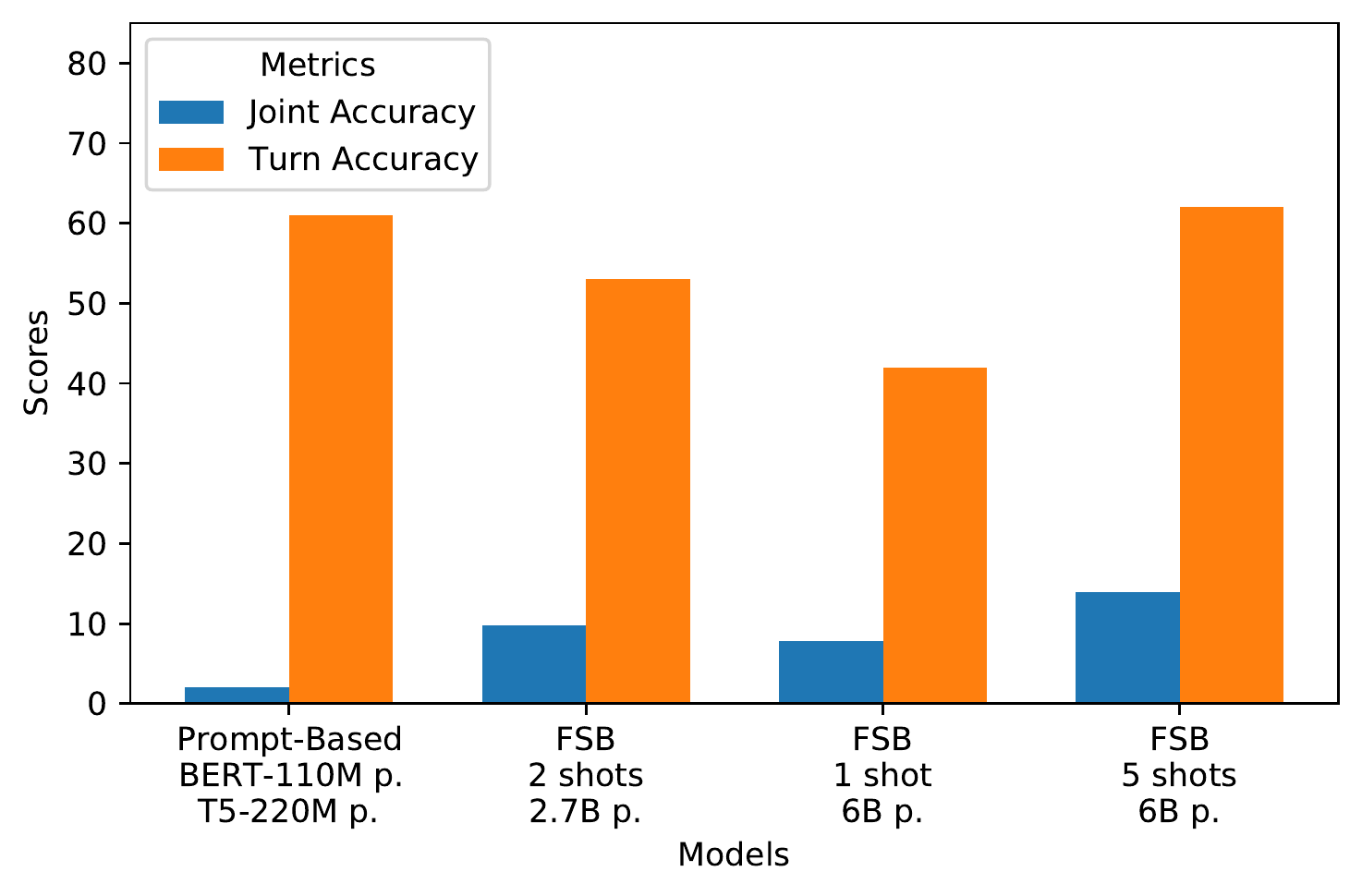}
        \caption{Few-Shot Baselines on 1-domain subset}
        \label{fig:fewmodel}
    \end{subfigure}
    \caption{Comparison with existing models}
\end{figure}

\noindent\textbf{Comparison with few-shot baselines:} The only model that presents results on few-shot belief state tracking for the MultiWoz dataset is the Few-Shot Bot \citep{madotto2021few}, which uses language models with knowledge retrievers for performing few-shot inference across multiple tasks. They evaluate their model on a single-domain subset of the MultiWoz dataset. We can see that our prompt-based approach, despite being trained on models having fewer parameters (330M v/s 6B) and on just 128 and 80 data points respectively, gets a lower joint accuracy (2\%) but a similar slot accuracy (61\%) as the best FSB baselines (Figure \ref{fig:fewmodel}).

\noindent\textbf{Comparison with finetuned models:} We compare with the end-to-end finetuned state-of-the-art models, GLAD \citep{zhong2018global}, SUMBT \citep{lee-etal-2019-sumbt}, TRADE \citep{wu2019transferable} and DSTQA \citep{zhou2019multi}. From Figure \ref{fig:model}, can get a sense of the large difference between these and this few-shot approach. 

\section{Conclusion and Future Directions}
This work highlights the difficulties in applying current prompt-based methods for dialog state tracking or, in general, complex multi-step tasks. We find several reasons for this.
First is the \emph{propagation of error} through the pipeline. A possible improvement can be to iteratively update the belief-state in each dialogue turn, similar to \citep{madotto2021few}. In this method, provided a proper correction mechanism, errors can be rectified in future dialogue turns. The second reason is the absence of appropriate prompts for domain prediction. This can benefit from the advent of ideas such as soft-prompting \cite{qin2021learning}.
Lastly, prompt-based models have a dependency on large pretrained models, which can lead to high inference delays. Our experiments show that although prompt-based learning has shown promising performance for few-shot classification problems, its application to more complex tasks is still an open problem and needs further research.
\bibliography{anthology,custom}
\bibliographystyle{acl_natbib}

\appendix

\section{Appendix}

\subsection{Prompts and formatted texts}
Here, we present all the prospective prompts ranked by their zero-shot performance for the domain-prediction task, where $DH$ is the dialogue-history.

\begin{itemize}[noitemsep,topsep=0pt,parsep=0pt,partopsep=0pt,leftmargin=*]
\item $DH$ Excited to see the $[MASK]$
\item $DH$ I am looking forward to see the $[MASK]$.
\item $DH$ Are any more detailts required pertaining to the $[MASK]$
\item Can you help me out about a $[MASK]$ $DH$
\item $DH$ Any other questions in regard to the $[MASK]$
\item I need some information about a $[MASK]$. $DH$
\item $DH$ So, we talked about the $[MASK]$ right?
\item $DH$ Thank you for helping me get all the information regarding the $[MASK]$.
\item I need some assistance in regards to finding a $[MASK]$ $DH$
\item $DH$ Thank you for the information on the $[MASK]$.
\item $DH$ So we are settled about the $[MASK]$ right?
\item I would need a $[MASK]$. $DH$
\item $DH$ I would need a $[MASK]$
\item $DH$ Services present are $[MASK]$.
\end{itemize}

\begin{figure}[h]
    \centering
    \includegraphics[width=\linewidth]{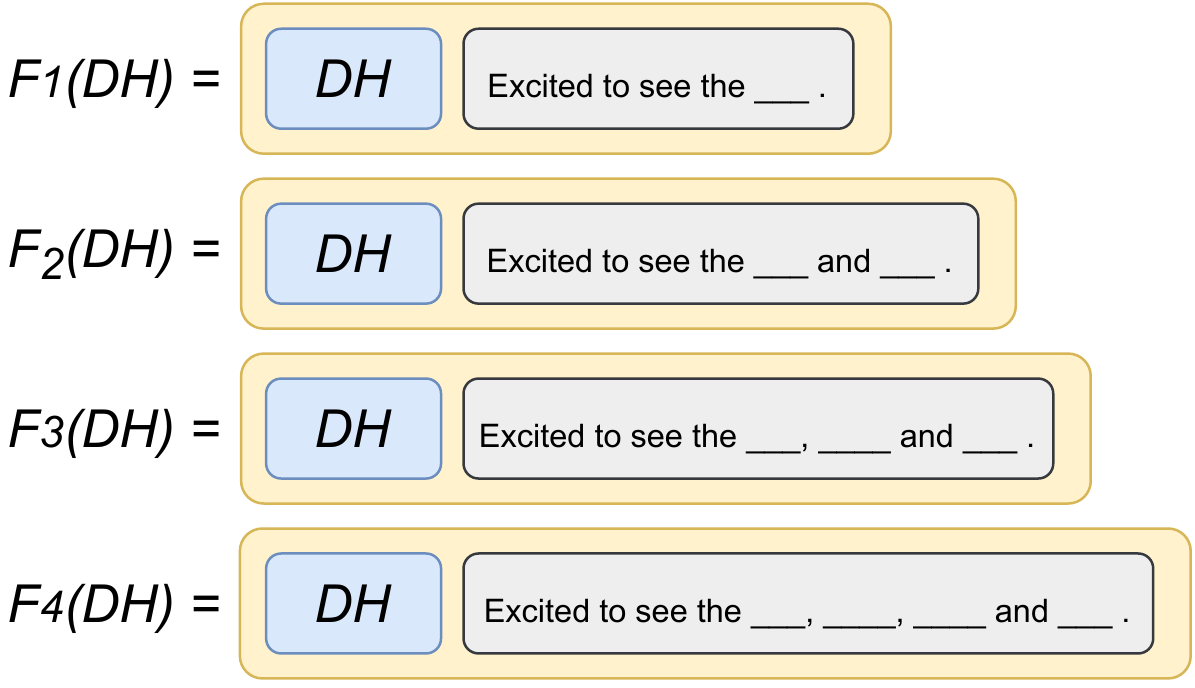}
    \caption{Variants of the adopted domain-prediction prompt concatenated to the Dialogue History ($DH$)}
    \label{fig:bertprompt}
\end{figure}

\begin{figure}[h]
    \centering
    \includegraphics[width=\linewidth]{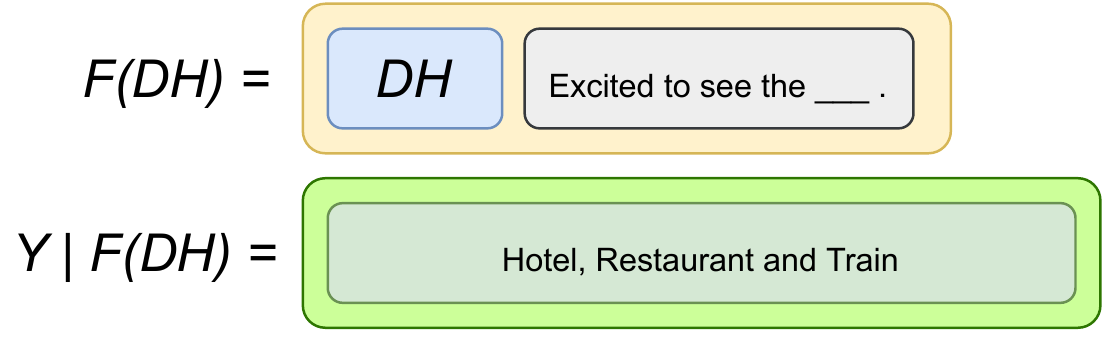}
    \caption{Formatted Dialogue History ($DH$) with QA-prompt and answer for slot-value prediction ($Y$)}
    \label{fig:gptprompt}
\end{figure}

\begin{figure}[!hbt]
    \centering
    \includegraphics[width=1\linewidth]{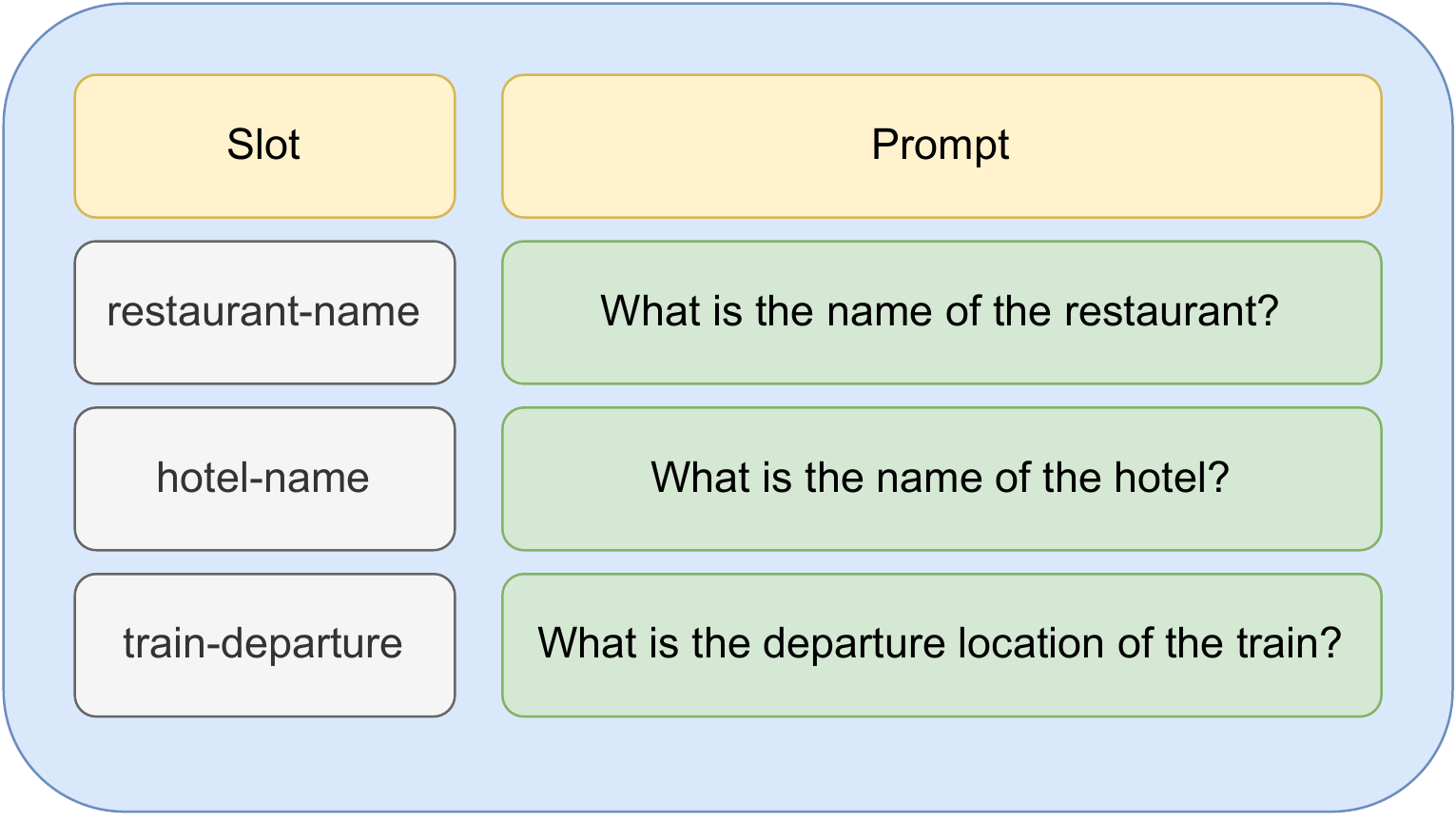}
    \caption{Slot-to-Prompt mapping examples}
    \label{fig:T}
\end{figure}

We provide examples of prompt-formatted inputs that are used during inference in Figures \ref{fig:bertprompt} and \ref{fig:gptprompt}. For slot-value prediction, we use the slot-dependant prompts as displayed in \ref{fig:T}.

\subsection{Detailed results and further analysis} \label{res}
Here, we provide the detailed results from each experiment in tables \ref{res1}, \ref{res2}, \ref{res3}, \ref{res4}.

\begin{figure}[!hbt]
    \centering
    \includegraphics[width=\linewidth]{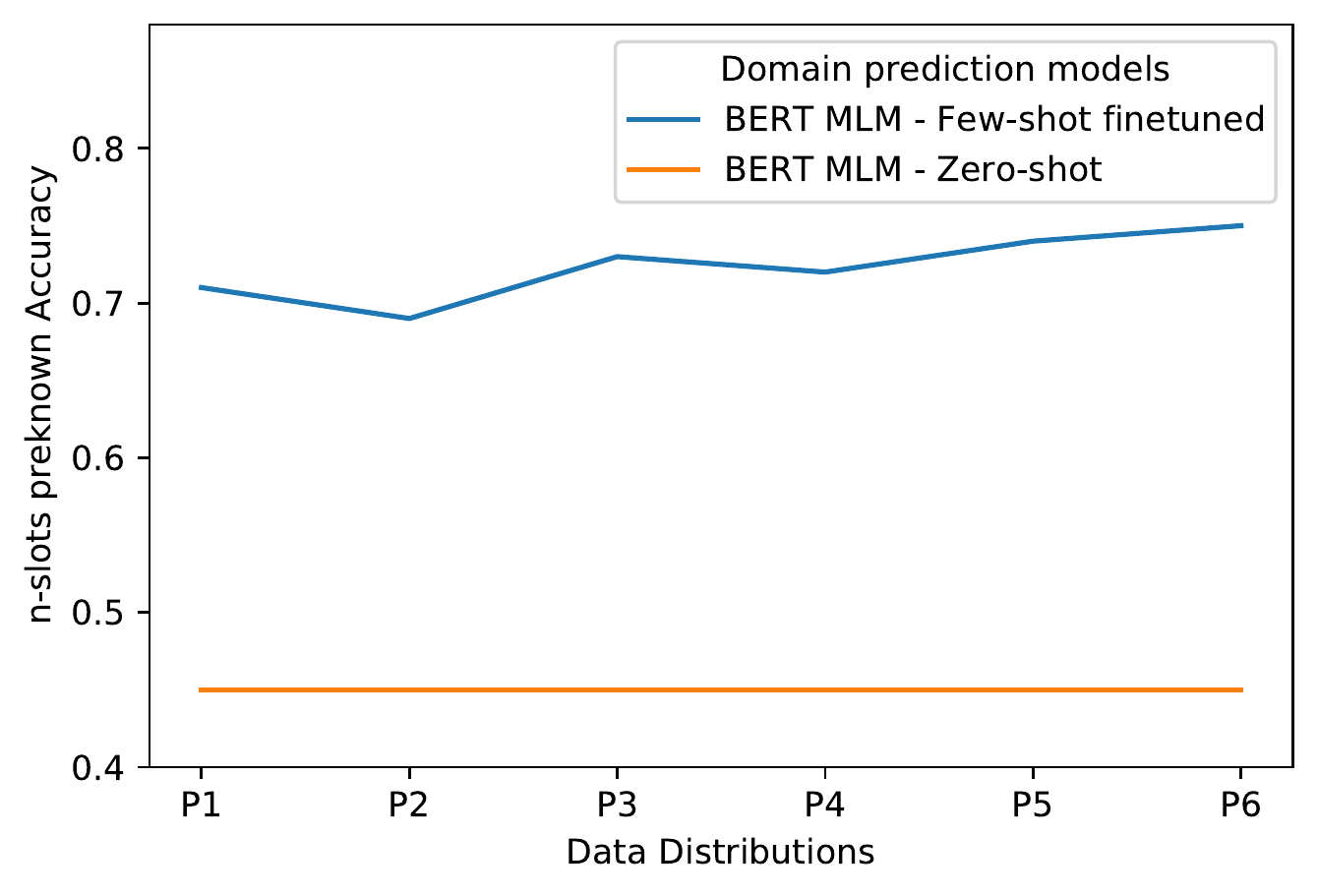}
    \caption{Domain prediction full accuracy variation with data distribution given the number of domains is known during inference for BERT-MLM}
    \label{fig:PK}
\end{figure}


\begin{table*}[]
\centering
\scriptsize
\begin{tabular}{|ccc|cc|}
\hline
\multicolumn{1}{|c|}{Task}                                                                                            & \multicolumn{1}{c|}{Model}                                                                                       & Loss Function             & \multicolumn{2}{c|}{Best Metrics across training runs}                                                                                                                 \\ \hline
\multicolumn{3}{|c|}{}                                                                                                                                                                                                                                               & \multicolumn{1}{c|}{Full Accuracy}                                                           & Partial Accuracy                                                        \\ \hline
\multicolumn{1}{|c|}{\multirow{3}{*}{\begin{tabular}[c]{@{}c@{}}Domain\\ Prediction\\ System (DP)\end{tabular}}}      & \multicolumn{1}{c|}{BERT MLM}                                                                                    & NLL Loss                  & \multicolumn{1}{c|}{0.47}                                                                    & 0.70                                                                    \\
\multicolumn{1}{|c|}{}                                                                                                & \multicolumn{1}{c|}{GPT2 LM}                                                                                     & NLL Loss                  & \multicolumn{1}{c|}{0.48}                                                                    & 0.68                                                                    \\
\multicolumn{1}{|c|}{}                                                                                                & \multicolumn{1}{c|}{GPT2 LM}                                                                                     & Unlikelihood Loss         & \multicolumn{1}{c|}{0.49}                                                                    & 0.69                                                                    \\ \hline
\multicolumn{3}{|c|}{}                                                                                                                                                                                                                                               & \multicolumn{1}{c|}{Accuracy}                                                                & Flexible Accuracy                                                       \\ \hline
\multicolumn{1}{|c|}{\multirow{3}{*}{\begin{tabular}[c]{@{}c@{}}Slot-value\\ Prediction\\ System (SvP)\end{tabular}}} & \multicolumn{1}{c|}{GPT2 LM}                                                                                     & \multirow{3}{*}{NLL Loss} & \multicolumn{1}{c|}{0.164}                                                                   & 0.180                                                                   \\
\multicolumn{1}{|c|}{}                                                                                                & \multicolumn{1}{c|}{T5 LM}                                                                                       &                           & \multicolumn{1}{c|}{0.343}                                                                   & 0.361                                                                   \\
\multicolumn{1}{|c|}{}                                                                                                & \multicolumn{1}{c|}{GPT-Neo LM}                                                                                  &                           & \multicolumn{1}{c|}{0.440}                                                                   & 0.446                                                                   \\ \hline
\multicolumn{3}{|c|}{}                                                                                                                                                                                                                                               & \multicolumn{1}{c|}{Turn Accuracy}                                                           & Joint Accuracy                                                          \\ \hline
\multicolumn{1}{|c|}{\multirow{2}{*}{\begin{tabular}[c]{@{}c@{}}Combined\\ System-1\\ (End-to-End)\end{tabular}}}     & \multicolumn{1}{c|}{\multirow{2}{*}{\begin{tabular}[c]{@{}c@{}}GPT2 LM (DP)\\ with\\ T5 LM (SvP)\end{tabular}}}  & \multirow{2}{*}{N.A.}     & \multicolumn{1}{c|}{\textbf{\begin{tabular}[c]{@{}c@{}}0.56\\ (Multi-domain)\end{tabular}}}  & \textbf{\begin{tabular}[c]{@{}c@{}}0.01\\ (Multi-domain)\end{tabular}}  \\ \cline{4-5} 
\multicolumn{1}{|c|}{}                                                                                                & \multicolumn{1}{c|}{}                                                                                            &                           & \multicolumn{1}{c|}{\begin{tabular}[c]{@{}c@{}}0.58\\ (Single-domain)\end{tabular}}          & \begin{tabular}[c]{@{}c@{}}0.02\\ (Single-domain)\end{tabular}          \\ \hline
\multicolumn{1}{|c|}{\multirow{2}{*}{\begin{tabular}[c]{@{}c@{}}Combined\\ System-2\\ (End-to-End)\end{tabular}}}     & \multicolumn{1}{c|}{\multirow{2}{*}{\begin{tabular}[c]{@{}c@{}}BERT MLM (DP)\\ with\\ T5 LM (SvP)\end{tabular}}} & \multirow{2}{*}{N.A}      & \multicolumn{1}{c|}{\begin{tabular}[c]{@{}c@{}}0.51\\ (Multi-domain)\end{tabular}}           & \begin{tabular}[c]{@{}c@{}}0.00\\ (Multi-domain)\end{tabular}           \\ \cline{4-5} 
\multicolumn{1}{|c|}{}                                                                                                & \multicolumn{1}{c|}{}                                                                                            &                           & \multicolumn{1}{c|}{\textbf{\begin{tabular}[c]{@{}c@{}}0.61\\ (Single-domain)\end{tabular}}} & \textbf{\begin{tabular}[c]{@{}c@{}}0.02\\ (Single-domain)\end{tabular}} \\ \hline
\end{tabular}
\caption{Best performing models across all sub-parts of the DST task: For the combined system, we prefer GPT2-LM domain prediction with T5-LM slot-value prediction. We observe that GPT-neo gives low final performance due to over-prediction of values. For the single-domain prediction case, we prefer BERT-MLM domain prediction, as we can constrain the prediction to only one domain using the appropriate prompt.}
\end{table*}

\noindent\textbf{Domain prediction with a known number of domains to predict:} One additional subject of interest is the full accuracy for BERT-MLM based domain prediction, when the number of domains to predict is known beforehand, thus, the prompt choice is predetermined (Figure \ref{fig:PK}). These results are significant because of the following reasons -
\begin{itemize}
\item With a pre known prompt, we get a full accuracy of upto 0.75, about 30\% more than the actual full accuracy, which goes on to show the potential of MLM-based domain predictions with proper output sampling.
\item  With an increasing proportion of data having up to 3 or 4 ground truth domains in the training-set, it shows an \textbf{increase in full accuracy}, as these data samples provide more data (in terms of masks to predict) per data sample and over-prediction of domains is not an issue given the pre-specified number of masks to include in the prompt.
\end{itemize}

\noindent\textbf{Comparisons of domain-prediction model performances with dataset size:}
In this section, we compare the performances for BERT-MLM and GPT-2 LM models for a range of dataset sizes, maintaining constant data distribution. Since there are no existing baselines for the multi-domain prediction task, we design a naïve \textbf{Keyword-based Classifier}. Here, we identify a few keywords manually from the dialogues that frequently come in conjunction with each domain. While predicting, we include a domain if any of the corresponding keywords appear in the dialogue history.

\begin{figure}[!hbt]
    \centering
    \includegraphics[width=\linewidth]{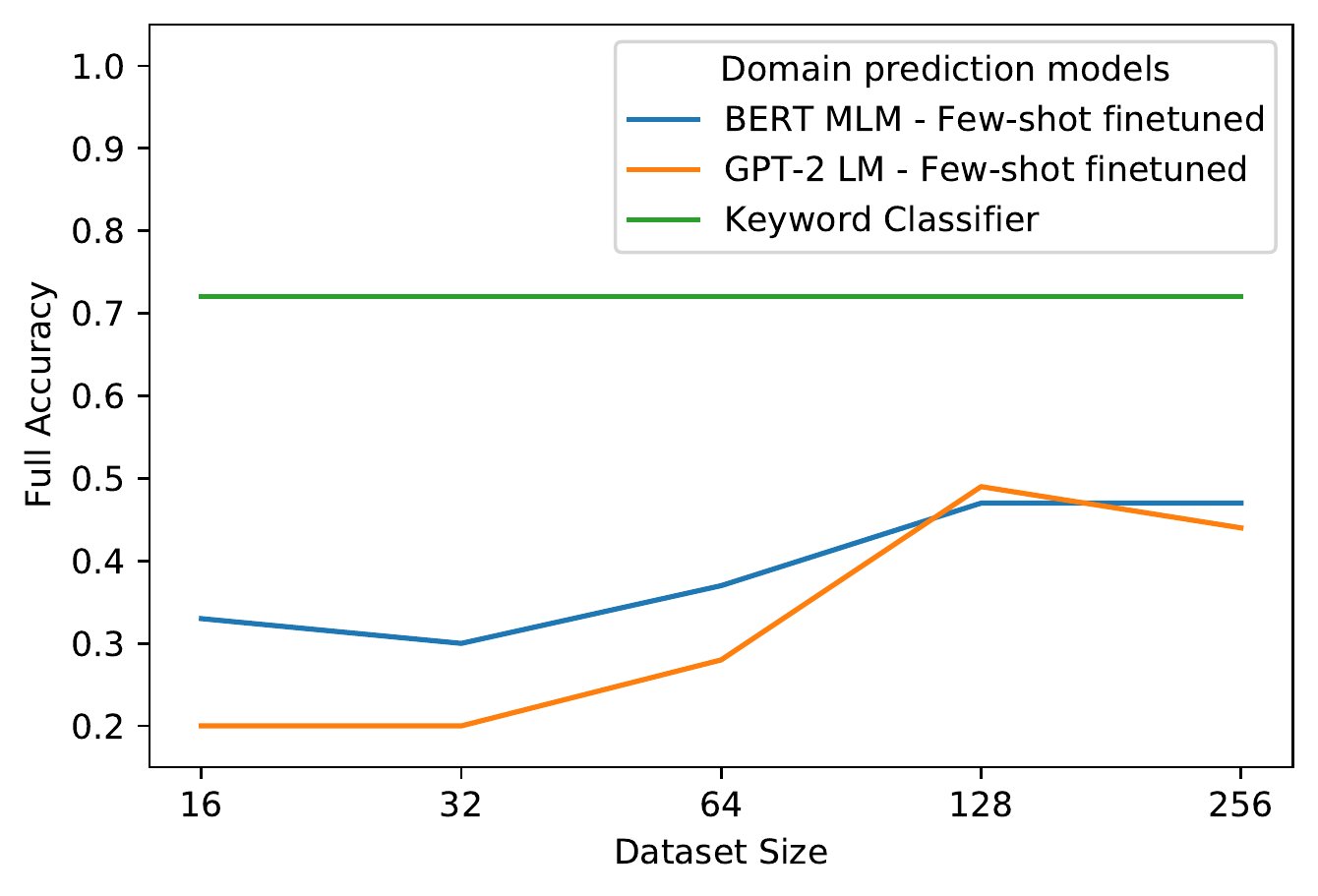}
    \caption{Full accuracy comparison of BERT-MLM and GPT-2 LM for different dataset sizes}
    \label{fig:FAcomp}
\end{figure}

\begin{figure*}[h]
    \centering
    \includegraphics[width=0.75\textwidth]{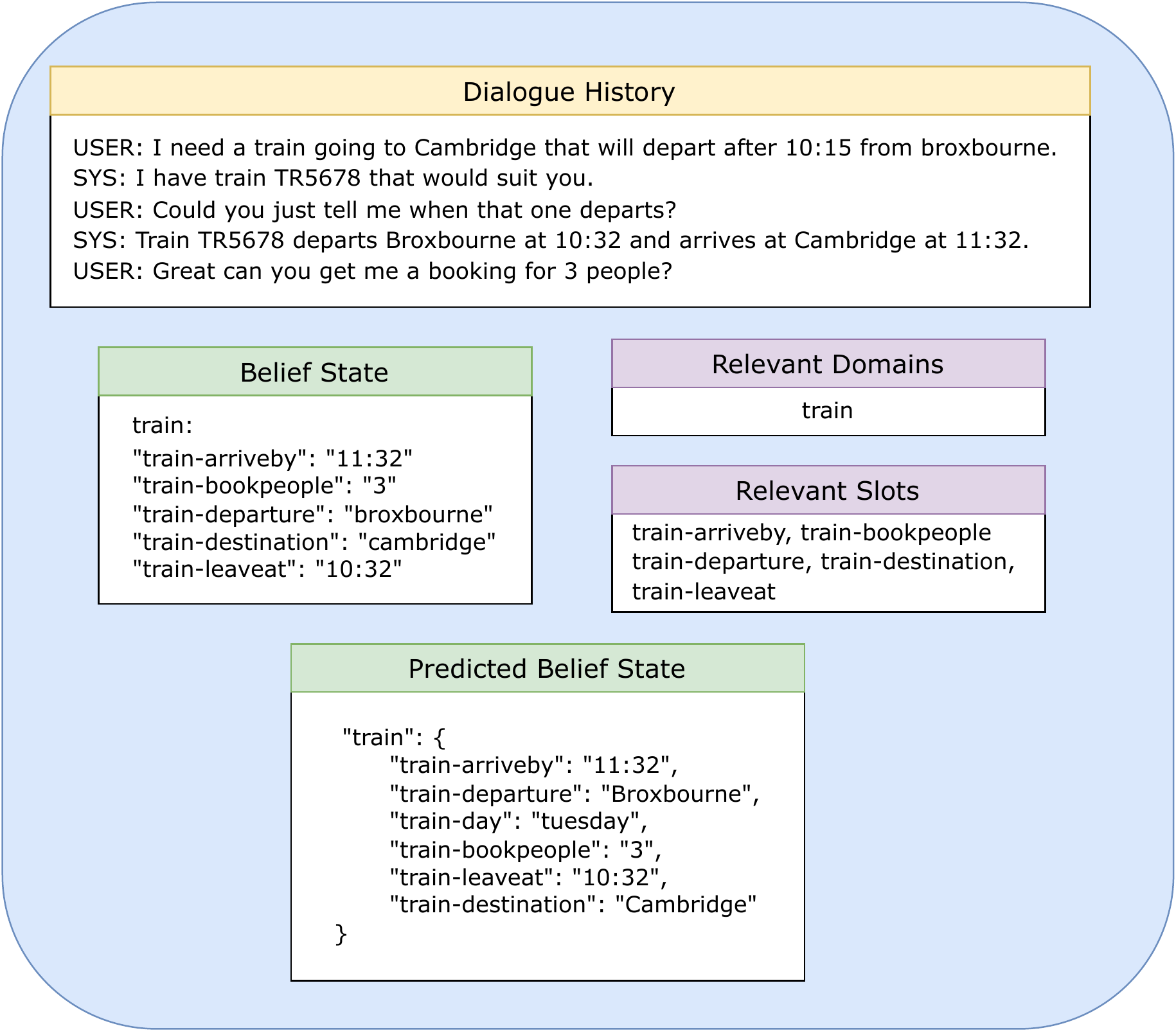}
    \caption{Example of MutiWOZ dialogue with ground-truth and predicted belief-states}
    \label{fig:exx}
\end{figure*}

According to Figure \ref{fig:FAcomp}, among the LM-based models, we can see that the GPT-2 accuracy is quite low when trained on smaller datasets, but rapidly increases with the dataset size. This is likely because of the less-constrained nature of generation adopted in GPT-2 LM.

The keyword-based classifier \emph{(Keyword Clf)} obtains a higher accuracy than the LM-based predictors. However, they make some unavoidable mistakes if the selected keywords are absent or are present in a different context. In general, these keyword-based approaches are neither scalable nor very robust, and hence are not preferable.

\noindent\textbf{Slot-value prediction for categorical slots:}
For slot-value prediction using categorical slots, the possible values that can be predicted are known beforehand (For example, any question involving a day of the week). In that case, it suffices to obtain the predictions by generating just one token and comparing with the first token for each of the possible slots, instead of generating the entire string. 

Assuming $X$ to be the dialogue history and $F_S(X)$, the dialogue history concatenated with the prompt corresponding to slot S, which is $p = T(S)$. Also assume $I: V \mapsto v_0$ to be the mapping from value to its first token and $J: v_0 \mapsto V$ as the inverse mapping. Given set of all possible slot-values $Vals$, the probability of the predicted slot to be $V$ is given as follows - 
$$
P(V|X, S) = \frac{P_M(Y_0 = I(V)\;|\;F_S(X))}{\sum_{V' \in Vals} P_M(Y_0 = I(V')\;|\;F_S(X))}
$$
Where $P_M$ is the language modelling score for some token. One key thing to note is that, this method only works when all the categorical slots have different first tokens, so the mappings are unique.

\paragraph{Implementation Details:} For masked domain-prediction, we used pretrained masked language model \textbf{BERT} \citep{devlin2018bert}. It was finetuned using Adam optimizer with a learning rate of 1e-7. The maximum number of epochs was capped to 20 and the batch-size was set to 8. Inference predictions were obtained with a single forward pass only. We use \textbf{GPT-2} \citep{brown2020language} for CLM-based domain prediction. GPT-2 was finetuned using Adam\citep{kingma2014adam} optimizer with a learning rate of 1e-7. The model was trained for 50 epochs, with a batch-size of 2. For less amount of data and a fixed number of epochs, we found that a batch-size of 2 gave better performance than a batch-size of 4 due to the more frequent weight updates in the former case. During inference, we used beam search decoding with beam size 5.

For the slot-value prediction task, we use \textbf{GPT-2} \cite{Radford2019LanguageMA}, \textbf{GPT-neo} \cite{brown2020language} and \textbf{T5} \cite{raffel2019exploring}.  
T5 was trained using Adam optimizer using a learning rate 1e-5 and the number of epochs was capped to 30. The GPT models were trained using the same hyper-parameters as used previously. During inference, we used beam search decoding with beam size 5. The pretrained model configs are given below -
\begin{itemize}[noitemsep,topsep=0pt,parsep=0pt,partopsep=0pt]
    \item \textbf{BERT}: \href{https://huggingface.co/bert-base-uncased}{\color{blue}{bert-base-uncased}}, 12-layer, 768-hidden, 12-heads, 110M parameters. Trained on lower-cased English text.
    \item \textbf{GPT-2}: \href{https://huggingface.co/gpt2}{\color{blue}{gpt2}}, 12-layer, 768-hidden, 12-heads, 117M parameters. Trained by OpenAI on a very large corpus of English data.
    \item \textbf{T5}: \href{https://huggingface.co/t5-base}{\color{blue}{t5-base}}, ~220M parameters with 12-layers, 768-hidden-state, 3072 feed-forward hidden-state, 12-heads, Trained on English text: the Colossal Clean Crawled Corpus (C4).
    \item \textbf{GPT-neo}: \href{https://huggingface.co/EleutherAI/gpt-neo-125M}{\color{blue}{gpt-neo-125M}}, 12-layer, 768-hidden, 12-heads, 125M parameters. Trained on the Pile, a large scale curated dataset created by EleutherAI.
\end{itemize}

\paragraph{Hardware specification and computational cost:} The models were trained on a single Tesla P100 12 GB GPU for about 30 minutes to 3 hours, depending on the model. Full training pipeline would require 4-6 hours.




\begin{table*}[]
\centering
\begin{tabular}{|cccc|ccc|}
\hline
\multicolumn{4}{|c|}{K-domain data = 128 * pk}                                            & \multicolumn{3}{c|}{BERT MLM}                                                       \\ \hline
\multicolumn{1}{|c|}{1}    & \multicolumn{1}{c|}{2}    & \multicolumn{1}{c|}{3}    & 4    & \multicolumn{1}{c|}{Full}          & \multicolumn{1}{c|}{Partial}       & Pre-known \\ \hline
\multicolumn{1}{|c|}{0}    & \multicolumn{1}{c|}{0}    & \multicolumn{1}{c|}{0}    & 0    & \multicolumn{1}{c|}{0.36}          & \multicolumn{1}{c|}{0.60}          & 0.45      \\
\multicolumn{1}{|c|}{0.4}  & \multicolumn{1}{c|}{0.3}  & \multicolumn{1}{c|}{0.2}  & 0.1  & \multicolumn{1}{c|}{\textbf{0.47}} & \multicolumn{1}{c|}{\textbf{0.70}} & 0.71      \\
\multicolumn{1}{|c|}{0.35} & \multicolumn{1}{c|}{0.35} & \multicolumn{1}{c|}{0.15} & 0.15 & \multicolumn{1}{c|}{\textbf{0.47}} & \multicolumn{1}{c|}{\textbf{0.70}} & 0.69      \\
\multicolumn{1}{|c|}{0.25} & \multicolumn{1}{c|}{0.25} & \multicolumn{1}{c|}{0.25} & 0.25 & \multicolumn{1}{c|}{0.43}          & \multicolumn{1}{c|}{0.68}          & 0.73      \\
\multicolumn{1}{|c|}{0.2}  & \multicolumn{1}{c|}{0.2}  & \multicolumn{1}{c|}{0.3}  & 0.3  & \multicolumn{1}{c|}{0.41}          & \multicolumn{1}{c|}{0.67}          & 0.72      \\
\multicolumn{1}{|c|}{0.15} & \multicolumn{1}{c|}{0.15} & \multicolumn{1}{c|}{0.35} & 0.35 & \multicolumn{1}{c|}{0.42}          & \multicolumn{1}{c|}{0.67}          & 0.74      \\
\multicolumn{1}{|c|}{0.1}  & \multicolumn{1}{c|}{0.2}  & \multicolumn{1}{c|}{0.3}  & 0.4  & \multicolumn{1}{c|}{0.33}          & \multicolumn{1}{c|}{0.63}          & 0.75      \\ \hline
\end{tabular}
\caption{Full and partial accuracy metrics for BERT-MLM domain prediction}
\label{res1}
\end{table*}

\begin{table*}[]
\centering
\begin{tabular}{|cccc|cc|cc|}
\hline
\multicolumn{4}{|c|}{K-domain data = 128 * pk}                                            & \multicolumn{2}{c|}{GPT2 LM}                       & \multicolumn{2}{c|}{GPT2 LM UL}                                   \\ \hline
\multicolumn{1}{|c|}{1}    & \multicolumn{1}{c|}{2}    & \multicolumn{1}{c|}{3}    & 4    & \multicolumn{1}{c|}{Full}          & Partial       & \multicolumn{1}{c|}{Full}          & \multicolumn{1}{l|}{Partial} \\ \hline
\multicolumn{1}{|c|}{0}    & \multicolumn{1}{c|}{0}    & \multicolumn{1}{c|}{0}    & 0    & \multicolumn{1}{c|}{0.21}          & 0.28          & \multicolumn{1}{c|}{-}             & -                            \\
\multicolumn{1}{|c|}{0.4}  & \multicolumn{1}{c|}{0.3}  & \multicolumn{1}{c|}{0.2}  & 0.1  & \multicolumn{1}{c|}{0.46}          & 0.66          & \multicolumn{1}{c|}{0.47}          & 0.67                         \\
\multicolumn{1}{|c|}{0.35} & \multicolumn{1}{c|}{0.35} & \multicolumn{1}{c|}{0.15} & 0.15 & \multicolumn{1}{c|}{0.46}          & 0.66          & \multicolumn{1}{c|}{\textbf{0.49}} & \textbf{0.69}                \\
\multicolumn{1}{|c|}{0.25} & \multicolumn{1}{c|}{0.25} & \multicolumn{1}{c|}{0.25} & 0.25 & \multicolumn{1}{c|}{\textbf{0.48}} & \textbf{0.68} & \multicolumn{1}{c|}{0.46}          & 0.67                         \\
\multicolumn{1}{|c|}{0.2}  & \multicolumn{1}{c|}{0.2}  & \multicolumn{1}{c|}{0.3}  & 0.3  & \multicolumn{1}{c|}{0.43}          & 0.64          & \multicolumn{1}{c|}{0.42}          & 0.65                         \\
\multicolumn{1}{|c|}{0.15} & \multicolumn{1}{c|}{0.15} & \multicolumn{1}{c|}{0.35} & 0.35 & \multicolumn{1}{c|}{0.41}          & 0.61          & \multicolumn{1}{c|}{0.35}          & 0.60                         \\
\multicolumn{1}{|c|}{0.1}  & \multicolumn{1}{c|}{0.2}  & \multicolumn{1}{c|}{0.3}  & 0.4  & \multicolumn{1}{c|}{0.36}          & 0.59          & \multicolumn{1}{c|}{0.31}          & 0.57                         \\ \hline
\end{tabular}
\caption{Full and partial accuracy metrics for GPT-2 LM domain prediction (UL = Unlikelihood Training)}
\label{res2}
\end{table*}

\begin{table*}[]
\centering
\begin{tabular}{|c|cc|cc|}
\hline
Data-Size & \multicolumn{2}{c|}{GPT2 LM UL}                    & \multicolumn{2}{c|}{BERT MLM}                      \\ \hline
          & \multicolumn{1}{c|}{Full Acc}      & Partial Acc   & \multicolumn{1}{c|}{Full Acc}      & Partial Acc   \\ \hline
16        & \multicolumn{1}{c|}{0.20}          & 0.28          & \multicolumn{1}{c|}{0.33}          & 0.60          \\
32        & \multicolumn{1}{c|}{0.20}          & 0.29          & \multicolumn{1}{c|}{0.30}          & 0.59          \\
64        & \multicolumn{1}{c|}{0.28}          & 0.45          & \multicolumn{1}{c|}{0.37}          & 0.64          \\
128       & \multicolumn{1}{c|}{\textbf{0.49}} & \textbf{0.69} & \multicolumn{1}{c|}{0.47}          & 0.70          \\
256       & \multicolumn{1}{c|}{0.44}          & 0.65          & \multicolumn{1}{c|}{\textbf{0.47}} & \textbf{0.73} \\ \hline
\end{tabular}
\caption{Variation of accuracy metrics for GPT-2 LM and BERT MLM for domain prediction with dataset size}
\label{res3}
\end{table*}

\begin{table*}[]
\centering
\begin{tabular}{|c|c|c|c|c|}
\hline
Model                 & Trained   & Data-per-slot (Total) & Strict Acc     & Flexible Acc   \\ \hline
\multirow{4}{*}{GPT2} & Zero-shot & 0                     & 0.025          & 0.048          \\
                      & Few-shot  & 3 (54)                & 0.062          & 0.086          \\
                      & Few-shot  & 5 (80)                & 0.109          & 0.132          \\
                      & Few-shot  & 7 (125)               & \textbf{0.164} & \textbf{0.180} \\ \hline
\multirow{4}{*}{T5}   & Zero-shot & 0                     & 0.288          & 0.311          \\
                      & Few-shot  & 3 (54)                & 0.300          & 0.314          \\
                      & Few-shot  & 5 (80)                & \textbf{0.343} & \textbf{0.361} \\
                      & Few-shot  & 7 (125)               & 0.256          & 0.259          \\ \hline
\multirow{4}{*}{GPT-neo} & Zero-shot & 0                     & 0.056          & 0.059          \\
                      & Few-shot  & 3 (54)                & 0.432          & 0.432          \\
                      & Few-shot  & 5 (80)                & 0.417          & 0.425          \\
                      & Few-shot  & 7 (125)               & \textbf{0.440} & \textbf{0.446} \\ \hline
\end{tabular}
\caption{Accuracy metrics for different training configurations of GPT2, GPT-neo and T5 for slot-value prediction.}
\label{res4}
\end{table*}

\end{document}